\documentclass[runningheads]{llncs}
\usepackage{graphicx}
\begin{document}
\title{Transcending the Annotation Bottleneck: AI-Powered Discovery in Biology and Medicine}
%
%\titlerunning{Abbreviated paper title}
% If the paper title is too long for the running head, you can set
% an abbreviated paper title here
%
\author{Soumick Chatterjee\inst{1,2}\orcidID{0000-0001-7594-1188}}
\authorrunning{Soumick Chatterjee}
% First names are abbreviated in the running head.
% If there are more than two authors, 'et al.' is used.
%
\institute{Human Technopole, Milan, Italy \and Faculty of Computer Science, Otto von Guericke University Magdeburg, Magdeburg, Germany
\\
\email{contact@soumick.com}}
\maketitle              % typeset the header of the contribution
\begin{abstract}
The dependence on expert annotation has long constituted the primary rate-limiting step in the application of artificial intelligence to biomedicine. While supervised learning drove the initial wave of clinical algorithms, a paradigm shift towards unsupervised and self-supervised learning (SSL) is currently unlocking the latent potential of biobank-scale datasets. By learning directly from the intrinsic structure of data - whether pixels in a magnetic resonance image (MRI), voxels in a volumetric scan, or tokens in a genomic sequence - these methods facilitate the discovery of novel phenotypes, the linkage of morphology to genetics, and the detection of anomalies without human bias. This article synthesises seminal and recent advances in "learning without labels," highlighting how unsupervised frameworks can derive heritable cardiac traits, predict spatial gene expression in histology, and detect pathologies with performance that rivals or exceeds supervised counterparts.

\keywords{Unsupervised Learning \and Medical Imaging \and Phenotype Discovery \and Anomaly Detection \and Genomics.}
\end{abstract}

\section{Introduction: The Annotation Bottleneck and the Unsupervised Solution}
For the past decade, the standard workflow in biomedical data analysis has necessitated the curation of datasets, the manual annotation of regions of interest (e.g., tumours, lesions, or anatomical structures), and the training of supervised models to replicate these human labels. While effective for specific, narrow tasks, this approach is fundamentally constrained by the scarcity of high-quality labels, the inherent bias of human knowledge, and the high cost of expert time. Furthermore, supervised approaches typically discard the vast majority of information contained within high-dimensional data, focusing only on features relevant to the pre-defined label.

To overcome these limitations, the field has increasingly turned to unsupervised and self-supervised learning. A common critique of these methods is that they sacrifice accuracy for flexibility. However, recent evidence suggests this trade-off is vanishing. In an investigation of voxel-wise segmentation for additive manufacturing, Iuso et al. \cite{iuso2024voxel} compared sophisticated supervised models (such as UNet++) against unsupervised VAE-based approaches for porosity detection. Remarkably, they found that unsupervised models, particularly when post-processed, could achieve performance metrics (Average Precision 0.830) that rivalled or even exceeded their supervised counterparts (Average Precision 0.751) in challenging testing scenarios. This finding challenges the orthodoxy that supervised learning is invariably superior, suggesting that for complex, highly variable targets, a model that comprehends the fundamental data distribution may be more robust than one trained to mimic a limited set of human labels.

These techniques learn robust representations by solving "pretext" tasks—such as contrasting similar views of an image or reconstructing masked portions of data—rather than predicting extrinsic labels. Seminal works in computer vision established the efficacy of this approach: SimCLR \cite{chen2020simple} demonstrated that contrastive learning could produce visual representations comparable to supervised methods, while DINO \cite{caron2021emerging} utilised self-distillation with Vision Transformers (ViT) to capture semantic segmentation properties without explicit supervision. In the medical domain, these principles were adapted to address data heterogeneity by Azizi et al. \cite{azizi2021big}, paving the way for "foundation models" capable of discovering biological signals that may elude human observers.

\section{Unsupervised Learning in Medical Imaging}
The application of unsupervised learning in medical imaging has matured from simple dimensionality reduction to complex tasks involving phenotype discovery, anomaly detection, and image registration.

\subsection{Phenotype Discovery and Genetic Linkage}
A principal advantage of data-driven discovery is the capacity to define quantitative phenotypes that bridge the gap between macroscopic imaging and microscopic genetics. In the realm of multimodal learning, Taleb et al. \cite{taleb2022contig} introduced the "ContIG" framework, demonstrating that self-supervised contrastive learning could effectively integrate medical imaging with genetic data to improve disease prediction.

Building upon this, Radhakrishnan et al. \cite{radhakrishnan2023cross} utilised cross-modal autoencoders to learn holistic representations of the cardiovascular state. Expanding this frontier, Ometto et al. \cite{ometto2024hundreds} recently developed a 3D diffusion autoencoder (3DDiffAE) to analyse temporal cardiac MRIs from the UK Biobank. Unlike traditional methods relying on fixed parameters such as ejection fraction, this unsupervised model learnt a "latent space" of 182 phenotypes describing complex cardiac wall motion and structure. Crucially, Ometto et al. demonstrated that these latent phenotypes shared a genetic architecture with established cardiac diseases, revealing 89 significant genomic loci.

This principle extends to the microscopic scale. In computational pathology, Cisternino et al. \cite{cisternino2024self} utilised self-supervised Vision Transformers (ViT) trained on over 1.7 million histology tiles from the Genotype-Tissue Expression (GTEx) project. Their model, RNAPath, utilises these self-supervised features to predict spatial RNA expression levels directly from H\&E-stained slides, effectively bridging the gap between tissue morphology and transcriptomics without the need for expensive spatial transcriptomics assays.

\subsection{Robust Anomaly Detection}
One of the most immediate clinical applications of learning without labels is anomaly detection—identifying pathology as a deviation from the normative distribution. StRegA \cite{chatterjee2022strega} addressed this in neuroimaging, a pipeline utilising a context-encoding Variational Autoencoder (VAE). By learning the distribution of healthy brain anatomy, StRegA identifies regions that the model cannot accurately reconstruct, successfully localising brain tumours and other anomalies without ever observing a labelled tumour during training.

Building upon these foundational VAE-based approaches, recent advancements have introduced more sophisticated generative models. Li et al. \cite{li2025scaleaware} introduced Scale-Aware Contrastive Reverse Distillation (SCAD), a novel framework that enhances anomaly detection by leveraging multi-scale feature representations and contrastive learning. SCAD addresses the challenge of scale variance in medical anomalies by distilling knowledge from a pre-trained teacher network to a student network in a reverse manner, effectively capturing anomalies across different resolutions. Furthermore, Bercea et al. \cite{bercea2025evaluating} critically evaluated normative representation learning in generative AI for robust anomaly detection in brain imaging. Their work highlights the importance of robust normative learning to ensure that generative models accurately capture the variability of healthy brain anatomy, thereby improving the sensitivity and specificity of anomaly detection. Addressing the limitations of standard diffusion models in preserving fine details, Beizaee et al. \cite{beizaee2025mad} introduced MAD-AD, a masked diffusion framework for unsupervised brain anomaly detection. By incorporating masking strategies into the diffusion process, MAD-AD effectively mitigates the issue of noise accumulation during image reconstruction, leading to more precise anomaly localisation compared to traditional diffusion-based methods.

The field continues to diversify with the emergence of State Space Models (SSMs) like Mamba, which have led to new architectures such as MAAT (Mamba Adaptive Anomaly Transformer) \cite{maat2025mamba}. This model efficiently captures long-range dependencies in physiological data, offering a computationally efficient alternative to traditional Transformers. Furthermore, foundation models are being adapted for this task; Seeböck et al. \cite{seebock2024anomaly} recently utilised self-supervised learning to guide segmentation in retinal OCT scans, effectively using anomaly detection as a weak supervision signal.

\subsection{Image Registration}
Deformable image registration has traditionally been computationally expensive. Unsupervised deep learning methods like VoxelMorph \cite{balakrishnan2019voxelmorph} learn to predict deformation fields by optimizing image similarity metrics, achieving state-of-the-art accuracy with significantly faster inference times. Building upon these foundations, MICDIR \cite{chatterjee2023micdir} introduced a multi-scale inverse-consistent framework incorporating a self-constructing graph latent. By explicitly encoding global dependencies and enforcing cycle consistency, MICDIR demonstrated statistically significant improvements over VoxelMorph in both intramodal and intermodal brain MRI registration tasks.

\section{Deciphering the Molecular Code}
Beyond imaging, unsupervised learning is revolutionising genomics and molecular biology by treating biological sequences as a "language" of life.

\subsection{Genomic Sequence Modelling}
Just as large language models learn the structure of text, genomic models can learn the grammar of regulatory elements and gene expression without explicit labels. Seminal works like DNABERT \cite{ji2021dnabert} applied the BERT architecture to k-mer sequences of DNA, demonstrating that attention mechanisms could capture global and local genomic context. More recently, the Nucleotide Transformer \cite{dallatorre2023nucleotide} scaled this approach to billions of parameters, training on multispecies genomes to predict molecular phenotypes and variant effects.

\subsection{Single-Cell Analysis}
The advent of single-cell RNA sequencing (scRNA-seq) has provided high-resolution views of cellular heterogeneity, but the data is inherently high-dimensional, sparse, and noisy. Deep generative models like scVI (Single-cell Variational Inference) \cite{lopez2018deep} use variational inference to approximate the underlying probability distributions of gene expression. By learning a low-dimensional latent representation of each cell, scVI can correct for batch effects, impute missing values, and cluster cell types without reliance on pre-defined markers.

\section{Clinical and Therapeutic Frontiers}
The utility of unsupervised learning extends into the translation of biological insights into clinical practice and therapeutic development.

% \subsection{Generative Chemistry and Drug Discovery}
% In drug discovery, the vastness of the chemical space makes supervised screening inefficient. Unsupervised generative models have emerged as a powerful tool to navigate this space. Automatic Chemical Design \cite{gomez2018automatic} introduced the use of VAEs to map discrete molecular structures into a continuous latent space, allowing researchers to optimise molecular properties via gradient-based search. Similarly, MolGAN \cite{decao2018molgan} leverages GANs operating directly on molecular graphs to generate novel small molecules with high validity and diversity.

\subsection{Computational Phenotyping from Electronic Health Records}
Electronic Health Records (EHR) contain rich, longitudinal data on patient health. Unsupervised learning allows for "computational phenotyping"—the discovery of clinical patterns without manual cohort definition. Inspired by natural language processing, models like BEHRT \cite{li2020behrt} treat patient medical histories as sequences of events and use Transformer architectures to learn robust patient representations. These self-supervised embeddings can predict future disease risks and stratify patients into novel subtypes, effectively enabling precision medicine at the population scale.

\section{Concluding Remarks}
The transition from supervised to unsupervised learning marks a decisive maturation in biomedical AI, effectively circumventing the "annotation bottleneck" that has long stifled progress. No longer compromising on accuracy, these self-supervised frameworks now rival supervised counterparts and drive genuine discovery—from defining novel cardiac phenotypes to decoding the genomic "language" of life. By leveraging the intrinsic structure of data, the field is moving towards a holistic view of biology where insights are derived from the data itself rather than human bias.

Future research must focus on the convergence of these modalities into unified "foundation models" capable of reasoning across imaging, genomics, and electronic health records simultaneously. Additionally, the exploration of computationally efficient architectures, such as State Space Models (e.g. Mamba), offers a promising avenue for modelling long-range biological dependencies that traditional Transformers struggle to capture. Ultimately, the priority remains bridging the gap between these high-dimensional latent representations and interpretable, clinically actionable biomarkers.

\bibliographystyle{splncs04}
\bibliography{mybibliography}
\end{document}